\documentclass{article}
\usepackage{spconf,amsmath,amssymb, graphicx, float, subfigure}

\title{Dual-BRANCH POLSAR IMAGE CLASSIFICATION BASED ON GRAPHMAE AND LOCAL FEATURE EXTRACTION}
\name{Yuchen Wang$^1$, Ziyi Guo$^1$, Haixia Bi$^1$, Danfeng Hong$^2$, and Chen Xu$^{3,4}$\thanks{This work is supported by the National Key R\&D Program of
China under Grant 2022YFA1003800, NSFC under Grant
42201394, and Qinchuangyuan Talent Program under grant QCYRCXM-2022-30. 
(\emph{Corresponding author: Haixia Bi}, haixia.bi@xjtu.edu.cn)
}}
\address{$^1$ School of Information and Communications Engineering, 
Xi'an Jiaotong University, \\Xi’an 710049, China\\
$^2$ Aerospace Information Research Institute, Key Laboratory of Digital Earth Science, 
\\Chinese Academy of Sciences, Beijing 100094, China\\
$^3$Department of Mathematics and Fundamental Research, 
Peng Cheng Laboratory,\\
Shenzhen 518055, China\\
$^4$School of Mathematics and Statistics,  
Xi’an Jiaotong University,
Xi’an 710049, China}

\begin{document}
%
\maketitle
\begin{abstract}

\begin{figure*}[t]
\centering
\includegraphics [width=0.65\textwidth]{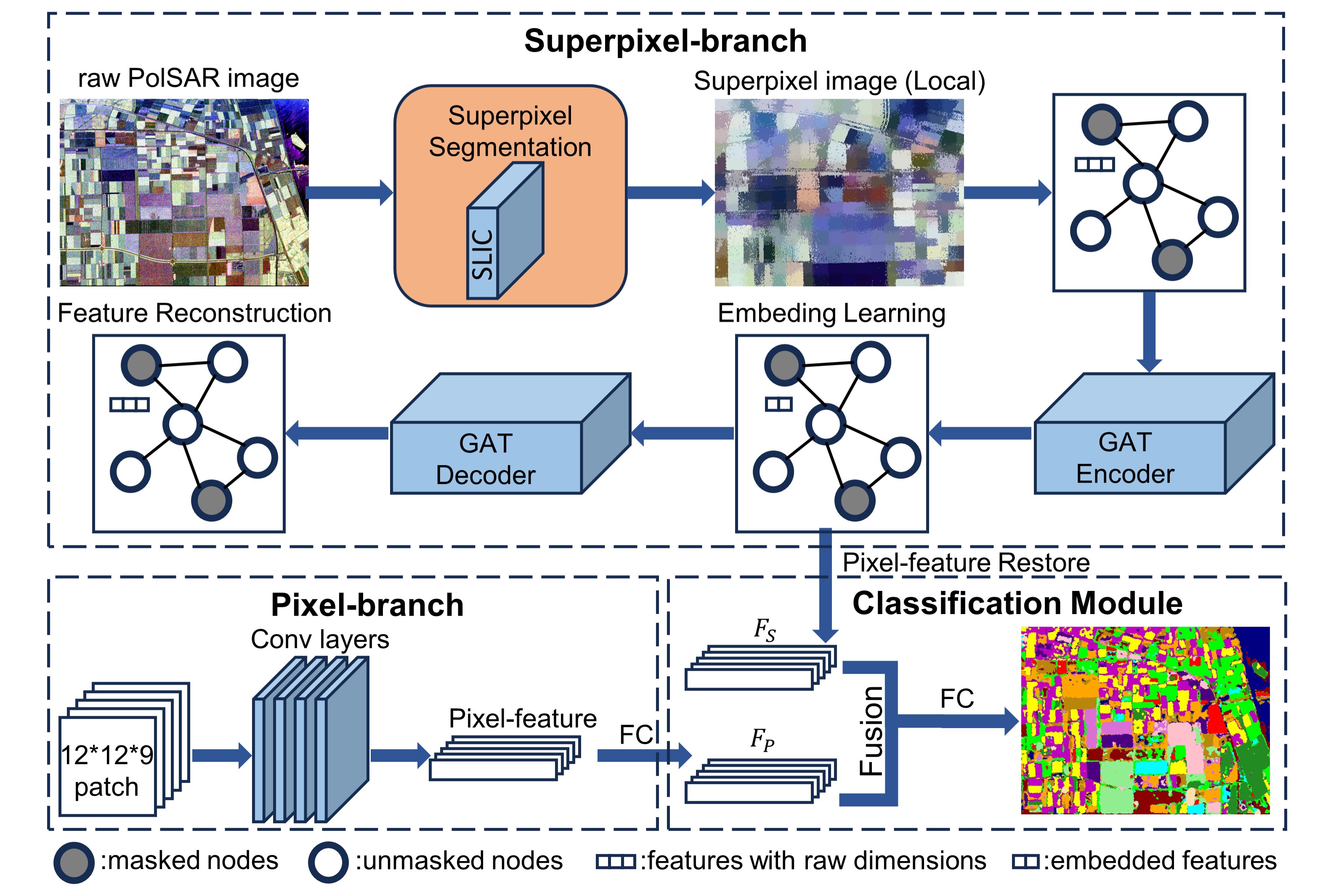}

\caption{Illustration the pipeline and architecture of our proposed model.}
\label{fig:example}
\end{figure*}

The annotation of polarimetric synthetic aperture radar (PolSAR) images is a labor-intensive and time-consuming process. 
Therefore, classifying PolSAR images 
with limited labels is a challenging 
task in remote sensing domain. 
In recent years, self-supervised learning approaches have proven effective in PolSAR image classification with sparse labels.
However, we observe a lack of research on generative self-supervised learning 
in the studied task. 
Motivated by this, we propose  
a dual-branch classification model 
based on generative self-supervised learning in this paper.
The first branch is a superpixel-branch, 
which learns superpixel-level polarimetric representations 
using a generative self-supervised graph masked autoencoder.
To acquire finer classification results, 
a convolutional neural networks-based 
pixel-branch is further incorporated to 
learn pixel-level features.
Classification with fused dual-branch features 
is finally performed to obtain 
the predictions.
Experimental results on 
the benchmark Flevoland dataset demonstrate 
that our approach yields promising classification results.
\end{abstract}
\section{Introduction}
\label{sec:intro}
PolSAR image classification endeavors to allocate real-world terrain classes to individual pixels, which is a pivotal task in remote sensing image processing realm. In recent years, machine learning algorithms have increasingly taken a dominant position, especially with the introduction of deep learning~\cite{wang2024towards,xiang2024polsar,bi2019unsupervised,bi2017unsupervised}. 
Notable examples, 
including real-valued convolutional neural networks (RV-CNN)~\cite{RV-CNN,bi2018graph,bi2019active,yang2023noise,bi2020polsar} and complex-valued convolutional neural networks (CV-CNN)~\cite{CV-CNN,wang2023complex,kuang2023complex}, 
remarkably enhanced PolSAR image classification performance. 
However, these deep learning-based algorithms usually require a large number of labels, 
posing great challenges to PolSAR image 
classification task, owing to the 
difficulty of obtaining a vast amount 
of labels.
Therefore, addressing PolSAR image classification with limited labels 
has currently become a key focus in remote sensing field.

Self-supervised learning (SSL)~\cite{ssl-survey}, 
which learns features without the aid of any manual labels, 
is a promising solution for the label scarcity issue. 
SSL can be categorized into two branches: 
contrastive learning and generative learning.
Contrastive learning aims to 
learn effective representations via 
distinguishing between positive 
and negative sample pairs.
For generative learning, the objective is to train an encoder to encode input $x$ into an embedding $z$ and a decoder to reconstruct $x$ from $z$~\cite{SSL}. 
Given the label-free nature of self-supervised learning, it aligns well with the 
label-hungry PolSAR image classification task. Recently, a number of SSL-based PolSAR image classification methods have been proposed~\cite{wang2023complex,kuang2023complex,TCSPANet,SSPRL}.
\cite{TCSPANet} 
proposes a two-staged contrastive learning based network for polarimetric representation 
learning, and designs a Transformer-based
sub-patch attention encoder to model the context within patch samples.
~\cite{kuang2023complex,SSPRL,10535284} put forward 
self-supervised PolSAR image classification methods  without the involvement of negative samples.

However, investigating existing methods, 
we observed a scarcity of researches in  
applying generative self-supervised learning 
approaches in PolSAR image classification.
Actually, generative self-supervised learning models 
have been demonstrated highly effective in representation learning, 
especially for classification tasks with limited labels~\cite{MAE,graphmae}.
Another advantage of generative models 
is that the model training does not rely on 
other techniques, such as positive or 
negative sample generation. 
Motivated by the above analysis, 
we explore the application of 
generative self-supervised learning 
in PolSAR image classification in this work.

Specifically, inspired by~\cite{graphmae}, 
we resort to graph masked autoencoder (GraphMAE) 
for feature learning, which 
achieves this goal by masking random nodes of the input graph 
and reconstructing the missing nodes with an encoder-decoder architecture.
In addition, the graph neural networks (GNN) backbone is able 
to capture the textual topology structure of images flexibly.
Within this framework, we model a given PolSAR image as an undirected graph, 
where the nodes correspond to pixels or patches, 
and the weighted edges represent similarities between the nodes. 
Considering that pixel-level topology structure of 
GNN often generates expensive computational costs~\cite{graph_cost}, 
a superpixel-level GraphMAE framework is designed in this work.
However, superpixel-level classification may result in coarse results, 
for the pixels within one superpixel are assigned with the same class label. 
Thereafter, to capture finer-grained features, a CNN-based 
pixel-branch is further integrated.

To sum up, we design a dual-branch PolSAR image classification approach in this paper, 
which aims to learn discriminative polarimetric representations 
with both GNN and CNN. Our contributions are summarized as follows:

1) To tackle the label scarcity issue in PolSAR image classification task, 
we propose to leverage generative graph-based self-supervised learning 
for polarimetric representation learning with an auxiliary graph reconstruction task,  
without the aid of any manual annotations.


2) To achieve fine-grained classification results, 
we further introduce a CNN-based pixel-level feature learning branch, 
which forms a dual-branch architecture for PolSAR image classification. 
This dual-branch design attains a balance 
between the model performance and computational costs. 
Experimental results on Flevoland benchmark dataset 
demonstrated the effectiveness of the proposed method.

The rest of this paper is organized as follows. 
Section 2 describes the proposed method. 
Section 3 presents the experiments and analysis. 
Section 4 finally concludes the paper.

\section{Method}
\label{sec:method}

Figure 1 illustrates the pipeline of our proposed model, which we call DB-GC 
(Dual-Branch with Graph masked autoencoder and Convolutional neural networks) 
in following sections. It consists of three components: 
a superpixel-branch based on graph masked autoencoder, 
a pixel-branch based on convolutional neural networks, 
and a classification module with fused features. 
Detailed explanations of these modules are provided in Section~\ref{S} to ~\ref{pred}.



\subsection{Superpixel-branch} \label{S}

In this section, we will first 
introduce the graph construction process and then 
explain the essential idea of the GraphMAE-based representation learning.



In this work, we use a 9-dimensional vector extracted from coherency matrix $T$ as the raw input feature:
\begin{equation}
\label{eq1}
\begin{array}{c}
[T_{11}, T_{22},T_{33},\mathbb{R}{(T_{12})},\mathbb{I}{(T_{12})}, \\
\quad \mathbb{R}{(T_{13})},\mathbb{I}{(T_{13})},
\mathbb{R}{(T_{23})},\mathbb{I}{(T_{23})}],
\end{array}
\end{equation}
where $\mathbb{R}(\cdot)$ and $\mathbb{I}(\cdot)$ 
denote the real and imagery component extracting operations respectively.

We first perform superpixel segmentation 
on the Pauli RGB image of a given PolSAR dataset  
with SLIC algorithm~\cite{SLIC}, 
generating a mask matrix $M_s$ indicating 
the pixels falling into each segment.
Given the superpixels, 
we next construct an undirected graph 
$G=\left\langle V,E,X\right\rangle$, 
where $V$ is the node set corresponding to superpixels,
$E$ denotes the edge set representing the similarities between adjacent nodes, 
and $X$ indicates features of nodes in $V$.
In this paper, the node feature is computed as 
the mean of the features over all pixels 
the superpixel contains.

With the constructed graph, 
we next sample a subset of nodes 
$\widetilde{V} \subset V$ and mask these nodes, 
which is implemented by replacing the node features 
with learnable vectors $x_{\left[M\right]}$.
Then the node feature $\widetilde{x_i}$ 
for $v_i \in V$ in the masked feature $\widetilde{V}$ 
can be expressed as:
\begin{equation}
\widetilde{x_i} = \begin{cases}
  x_{\left[M\right]} & \text{if } v_i \in \widetilde{V} \\
  x_i & \text{if } v_i \notin \widetilde{V}\text{. }
\end{cases}
\end{equation}

The objective of GraphMAE is to recover the masked 
node features using partially known nodes 
and edges $E$. 
The masked graph is input into an encoder implemented with graph attention network (GAT)~\cite{gae} to generate hidden representations $e_i$. 
Subsequently, the same nodes in $\widetilde{V}$ are remasked to $e_{\left[M\right]}$ 
and fed into a GAT-based decoder, 
reconstructing a graph which has 
the same structure as the input graph. 
The loss function of the model is defined as
\begin{equation}
L_{SCE} = \frac{1}{\left| \widetilde{V} \right|} \sum_{v_i \in \widetilde{V}} \left(1 - \frac{x_i^T z_i}{\left\| x_i \right\| \cdot \left\| z_i \right\|}\right)^\gamma\text{, }
\end{equation}
where $x_i$, $z_i$ denote the raw and reconstructed feature for node $v_i$ respectively, and $\gamma$ is a hyperparameter that scales the cosine error.

With the generative self-supervised learning process, 
$e_i$ works as the encoder for the downstream classification task. 
It should be noted that,
since the GraphMAE is employed in superpixel-level 
in this work, for each pixel in a superpixel, 
its feature takes the values of 
the superpixel feature.
We denote the pixel features as $F_s$ 
in the following.


\subsection{Pixel-branch} \label{P}

As the superpixel-branch targets at superpixel-level feature extraction, it encounters the limitation of being unable to distinguish 
between  pixels with different classes within the same superpixel. 
To address this issue, we introduce another branch, 
i.e., the pixel-branch, with the objective of learning fine-grained pixel-level features.


\begin{figure*}[t]
\centering
\includegraphics [width=0.70\textwidth]{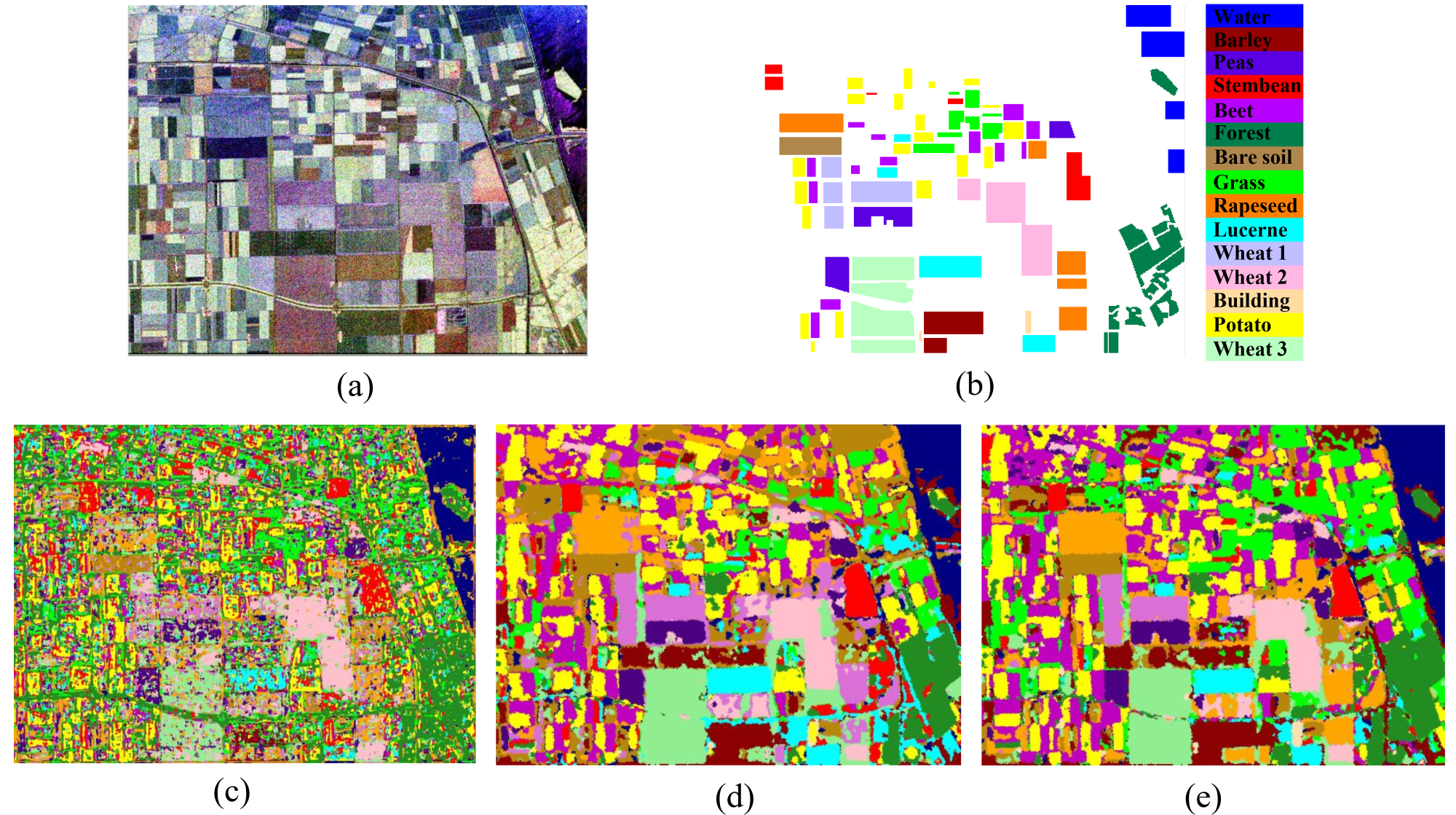}

\caption{Experimental data and results. (a) PauliRGB image. (b) Ground-truth. (c) CNN. (d) GNN. (e) DB-GC.}
\label{fig:exp}
\end{figure*}

For the pixel-branch, 
pixel-centered patches with size $n \times n$ 
are used as input. 
In this branch, we utilize 
CNN as the deep learning architecture.
The CNN consists of four blocks, 
where each block includes a convolutional layer 
and a ReLU activation function. 
In addition, each of the second and fourth layers 
has a max-pooling layer. 
The output channels for the four convolutional layers 
are 128, 256, 512 and 512, respectively. 
Following this, a fully-connected layer 
is employed for feature output. 
Finally, pixel-level feature set $F_p$ is 
obtained, which has the same dimension as $F_s$.

\subsection{Classification with fused features} \label{pred}

We next fuse the features extracted 
from the above two branches. 
Specifically, $F_s$ and $F_p$ are added together 
with weights $\alpha$ and $1-\alpha$ 
to obtain the final features $F$ as follows:
\begin{equation}
F = \alpha F_S +(1-\alpha)F_P\text{. }
\end{equation}

With the fused feature $F$, 
a fully-connected layer and a softmax layer 
are applied as the classifier.
Cross-entropy loss function~\cite{zhang2018generalized} 
is exploited as the optimization objective for the 
classifier training.

\section{EXPERIMENT}
\label{sec:experiement}

We evaluate the performance of the proposed model on the benchmark Flevoland dataset. 
The size of the image is 1024 × 750, 
with a total of 15 classes including water, barley, peas, stem beans, beet, forest, bare soil, grass, rapeseed, lucerne, potatoes, three types of wheat, and buildings.

\begin{table}[ht]
  \begin{center}

    \begin{tabular}{c|c|c|c} 
    \hline\hline
      \textbf{Class}  & \textbf{GNN} & \textbf{CNN} & \textbf{DB-GC}\\
      \hline
      Water & 0.9456 & 0.9779 & 0.9875\\
      Barley & 0.9744 & 0.3621 & 0.9863\\
      Peas & 0.9851 & 0.6351 & 1.0000\\
      Stembean & 0.9428 & 0.8319 & 0.9984\\
      Beet & 0.8610 & 0.5172 & 0.9721\\
      Forest & 0.9811 & 0.8164 & 0.9930\\
      Bare soil & 0.9807 & 0.7635 & 0.9636\\
      Grass & 0.6562 & 0.6781 & 0.9647\\
      Rapeseed & 0.3690 & 0.5319 & 0.9711\\
      Lucerne & 0.9871 & 0.5407 & 0.9926\\
      Wheat 1 & 0.9807 & 0.3912 & 0.9651\\
      Wheat 2 & 0.9616 & 0.8134 & 0.9888\\
      Building & 0.3355 & 0.0000 & 0.9435\\
      Potato & 0.8760 & 0.6908 & 0.9825\\
      Wheat 3 & 0.9943 & 0.6341 & 0.9905\\
      \hline
      OA & 0.8915 & 0.6669 & 0.9840\\
      AA & 0.8554 & 0.6123 & 0.9800\\
    \hline\hline
    \end{tabular}
    \caption{Numerical experimental results.}
  \end{center}
\end{table}

In the superpixel segmentation process, 
we divide the dataset into approximately 120,000 superpixels.
An equal number of pixels are selected as training set 
for each class, specifically 111 pixels per class, accounting for 1\% of all labeled pixels. 
 GraphMAE has 4 layers of encoder and 1 layer of decoder, and each layer is GAT with 4 heads. $~\gamma$ is set to 3 in the loss function.
 We set $~\alpha$ as 0.4. 
The training epoch numbers for 
the self-supervised training and 
joint dual-branch training are 400 
and 250 respectively.
Overall accuracy (OA) and average accuracy (AA) are 
used as the evaluation criteria.

To validate the effectiveness of our dual-branch design, 
we compared DB-GC with single-branch architectures  
which are implemented by setting $\alpha$ to 0 (CNN) or 1 (GNN). 
The results are presented in Table 1 and Fig. 2.
From these results, we can draw the following conclusions:

1) The visual results of the CNN model exhibit a considerable amount of speckle, which may be attributed to the CNN's excessive focus on local details and the scarcity of labeled data involved in training. In contrast, the GNN model demonstrates significantly better performance on both metrics and visual representations, indicating its capability to extract features from unlabeled data, no longer constrained by limited labeled data.

2) Compared to the standalone GNN, DB-GC model demonstrates increases in OA and AA scores by 9.25\% and 12.46\% respectively. Additionally, its visual results exhibit clearer boundaries, indicating that the incorporation of the CNN local feature extraction module successfully rectified the differences between pixels within the same superpixel.

3) The DB-GC model outperforms the other two single models in terms of accuracy for each class, with all accuracies exceeding 94\%. This demonstrates its ability to extract features through self-supervised learning and to learn local features effectively, particularly in datasets with limited data.





\section{CONCLUSION AND FUTURE WORK}
\label{sec:conclusion}

In this study, we propose a dual-branch PolSAR image classification model 
based on graph masked autoencoder and convolutional neural networks. 
The superpixel-branch learns superpixel-level polarimetric representations 
via a generative self-supervised graph masked autoencoder framework.
A convolutional neural networks-based pixel-branch
is further integrated to learn fine-grained pixel-wise features.
Predictions are finally made by fusing the dual-branch features.
Experiments on the benchmark Flevoland dataset 
demonstrated the effectiveness of the proposed approach.

In the future, considering the complex value characteristics of PolSAR data, 
we plan to design PolSAR image classification method 
based on complex-valued deep architectures, to 
make full use of the PolSAR information.


\bibliographystyle{IEEEbib}
\bibliography{refs}

\end{document}